\documentclass[review]{elsarticle}

\usepackage{microtype}
\usepackage{inconsolata}
\usepackage{multirow}
\usepackage{subfigure}
\usepackage{hyperref}
\usepackage{synttree}

\journal{Knowledge Based Systems}

\biboptions{authoryear}

\begin{document}

\begin{frontmatter}

\title{A Modular Approach for Multilingual Timex Detection and Normalization using Deep Learning and Grammar-based methods}
\author[myaddress]{Nayla Escribano}
\ead{nayla.escribano@ehu.eus}

\author[myaddress]{German Rigau}
\ead{german.rigau@ehu.eus}

\author[myaddress]{Rodrigo Agerri}
\ead{rodrigo.agerri@ehu.eus}

\address[myaddress]{HiTZ Center - Ixa University of the Basque Country UPV/EHU}




\begin{abstract}
Detecting and normalizing temporal expressions is an essential step for many
NLP tasks. While a variety of methods have been proposed for detection, best
normalization approaches rely on hand-crafted rules. Furthermore, most of them
have been designed only for English. In this paper we present a modular
multilingual temporal processing system combining a fine-tuned Masked Language
Model for detection, and a grammar-based normalizer. We experiment in Spanish
and English and compare with HeidelTime, the state-of-the-art in multilingual
temporal processing. We obtain best results in gold timex normalization, timex
detection and type recognition, and competitive performance in the combined
TempEval-3 relaxed value metric. A detailed error analysis shows that detecting
only those timexes for which it is feasible to provide a normalization is
highly beneficial in this last metric. This raises the question of which is the
best strategy for timex processing, namely, leaving undetected those timexes
for which is not easy to provide normalization rules or aiming for high
coverage.
\end{abstract}

\begin{keyword}
Temporal Processing \sep Multilingualism \sep Sequence Labelling \sep Grammar-based approaches \sep
Deep Learning \sep Natural Language Processing
\end{keyword}

\end{frontmatter}

\section{Introduction}\label{sec:intro}



Many Natural Language Processing (NLP) processes involve the extraction of
temporal information from large amounts of texts. This kind of information can
be used in a range of tasks, such as question-answering, to provide information
about when an event happened \citep{llorens15QAtempeval}; text summarization, to
identify and order different events to generate the representation of a text
\citep{aslam15summarization}; or information extraction, to find clinically
relevant temporally-anchored events in patient records
\citep{bethard17clinical}.

The TempEval-3 shared task \citep{uzzaman13te3} is a widely used evaluation
benchmark for the combined detection and normalization of timexes for English
and Spanish. For instance, in the fragment shown in Example (1) the timex
\textit{dos días} should be detected, classified as ``DURATION'' and normalized
as ``P2D'' (for a two-day period).

\begin{enumerate}
    \item[(1)] ... y se iniciará \textit{dos días} después de... \\
    (... and it will begin \textit{two days} after...)
\end{enumerate}

Timex extraction systems have traditionally used rule-based methods for both
detection and normalization processes
\citep{stroetgen13heidelTE3,chang13sutime,zhong17syntime}, or hybrid setups,
applying both supervised techniques and rules
\citep{llorens10tipsem,bethard13cleartk,lee14uwtime,ning18cogcomptime}. In the last years, new
approaches based on Transformer Masked Language Models (MLMs) have been
proposed, although mainly for the detection component
\citep{lange20multi,almasian21bert}. Systems that automate rule generation have also been proposed \citep{stroetgen15heidelmulti,ding21artime}. While \citet{laparra18semeval} propose an
alternative timex processing evaluation on the SCATE schema
\citep{bethard16scate}, TimeML \citep{pustejovsky05timeml} and TempEval-3 remains
the de-facto benchmark on temporal processing. Furthermore, most approaches are
only focused on English, with the exception of very few proposals
\citep{stroetgen13heidelTE3,stroetgen15heidelmulti,lange20multi}.

HeidelTime \citep{stroetgen13heidelTE3} is considered to be the
state-of-the-art in multilingual timex detection and normalization. They use a
monolithic approach in which rules, detection patterns and normalization values
must be all hand-crafted in an integrated manner, and efforts to automate this
rule-construction process have not led to comparable results in multilingual settings
\citep{stroetgen15heidelmulti}. In contrast, we propose a modular system for
timex detection and normalization, combining neural and rule-based methods as
the best techniques for each subtask and obtaining state-of-the-art results at
both tasks. This allows us to avoid any handcrafting of detection patterns and
to focus mostly on the normalization rules, thus streamlining the overall
development, also for the inclusion of new languages. Unlike HeidelTime, the
grammar we employ does not process context to disambiguate normalization values
and works from a general domain-blind perspective.

More specifically, we present a multilingual timex processing system for
English and Spanish. Our modular approach consists of a XLM-RoBERTa MLM
\citep{conneau20xlmroberta} fine-tuned for timex detection and classification
for both English and Spanish together, plus a TimeNorm SCFG rule-based
normalizer \citep{bethard13scfg} with independent English and Spanish grammars.
Our system outperforms HeidelTime at gold timex normalization and timex
detection in both languages and obtains similar results when evaluating the
TempEval-3 combination of the timex detection and normalization task. A
detailed error analysis suggests that jointly considering both detection and
normalization to only normalize those timexes that are correctly detected is
particularly beneficial in this last evaluation setting. 

The rest of the contributions are the following: (i) We present the first
TimeNorm grammar for Spanish which obtains state-of-the-art results in gold
timex normalization, an evaluation setting proposed by \citet{bethard13scfg};
(ii) we show how to develop a simple but effective modular system for
multilingual timex processing with minimal manual intervention; (iii) unlike
popular approaches such as Heideltime, our approach allows to pick and choose
the desired approach for each of the tasks involved, namely, detection and
normalization; (iv) we develop a new TempEval-3 compatible evaluator that works
with text in tabulated format, thereby facilitating the evaluation of sequence
labeling systems; (v) the Spanish grammar\footnote{\url{https://github.com/NGEscribano/timenorm-es}} and any developed code\footnote{\url{https://github.com/NGEscribano/XTN-timexes}} is publicly
available to facilitate research on temporal processing and for reproducibility
of results.

Related work on timex processing is presented in Section \ref{sec:related}.
Section \ref{sec:setup} describes the system setup and the resources that were
used, as well as the approach to build our Spanish grammar. Section
\ref{sec:results} explains the experiments that have been performed and the
results of our system compared to the best multilingual system that performs
both temporal detection and normalization, namely, HeidelTime.
Finally, a discussion on these results and the errors made by each approach is
presented in Section \ref{sec:discussion}.

\section{Related Work}\label{sec:related}

HeidelTime \citep{stroetgen13heidelTE3} is perhaps the most popular system for
timex processing. It uses a monolithic approach based on hand-crafted patterns
and rules for each language and domain. They also extend their resources
automatically to more than 200 languages \citep{stroetgen15heidelmulti},
although the results are significantly lower than for the manually developed
languages. Other rule-based systems such as SUTime \citep{chang13sutime} address
the task by means of a cascade of finite automata on regular expressions over
tokens and is part of the Stanford CoreNLP tools \citep{manning14corenlp}.
Moreover, SynTime \citep{zhong17syntime} benefits from the similar syntactic
behaviour of time expressions to define and find different token types such as
``text tokens'', ``modifiers'' and ``numerals'', but is only meant for timex
detection.

Hybrid systems combine rule-based and machine learning approaches. ClearTK
\citep{bethard13cleartk} uses a minimal set of simple morpho-syntactic features
to perform timex detection and the TimeN system for normalization
\citep{llorens12timen}, which is based on hand-crafted rules and can be applied
to different languages. TIPSem \citep{llorens10tipsem} is one of the few systems
that are not designed only for English, also addressing Spanish. They train a
CRF model \citep{lafferty01crf} using morpho-syntactic and semantic features to
find timexes and a set of rules to normalize them. UWTime \citep{lee14uwtime}
trains a context-dependent semantic parser for English based on a Combinatory
Categorial Grammar, learning from possible compositional meaning
representations to detect and normalize time expressions. CogCompTime \citep{ning18cogcomptime} provides a standalone temporal processing system that mixes machine learning techniques for timex detection and rules for normalization. Finally, ARTime \citep{ding21artime} uses an algorithm to automate the generation of normalization rules and combines this process with existing detection systems.

Recent approaches have mostly focused on neural and deep learning
techniques. \citet{lange20multi} train a bi-directional LSTM
\citep{hochreiter97lstm} with a CRF \citep{lafferty01crf} output layer on
pre-trained FastText \citep{bojanowski17fasttext} and multilingual BERT
embeddings \citep{devlin19bert}. They align embedding spaces to create a single
multilingual model using adversarial training, and apply their models to seen
and unseen languages. \citet{almasian21bert} experiment with variations of
token classification and sequence-to-sequence models based on BERT
\citep{devlin19bert} and RoBERTa \citep{liu19roberta}. Additional training on
weakly labeled data from HeidelTime is used to alleviate the high demand of
data of sequence-to-sequence models. On the other hand,
\citet{laparra18neural-norm} present a neural system for both timex detection
and normalization on the SCATE schema \citep{bethard16scate}, which considers
timexes as semantic compositions of time entities. They train different RNN
models using bi-directional GRUs on words and characters, finding that a
character-based multi-output neural architecture is better for timex processing
on SCATE.

Apart from \citet{laparra18neural-norm}, the majority of recent neural
approaches have focused on timex detection, while timex normalization
state-of-the-art results are still obtained by traditional rule-based and
hybrid systems \citep{bethard13scfg,stroetgen13heidelTE3,lee14uwtime}. On the
other hand, most systems have only addressed timex processing for English, with
very few exceptions
\citep{llorens10tipsem,stroetgen13heidelTE3,stroetgen15heidelmulti}. UWTime obtains the best performance for English, whereas HeidelTime remains as the best multilingual approach. Given the considerable effort to build a rule-based or hybrid system that achieves competitive results in a multilingual setting, we propose the combination of the best approaches for each
subtask to process timexes in different languages, namely, MLMs for detection and
rule-based techniques for normalization.


\section{Our Method}\label{sec:setup}

Our modular approach consists of two separate components: the timex detector and
the timex normalizer. The \textbf{timex detector} is a sequence labelling model
based on XLM-RoBERTa \cite{conneau20xlmroberta} fine-tuned to extract English
and Spanish time expressions and their types: TIME, DATE, DURATION and SET. The
\textbf{timex normalizer} is built on the TimeNorm SCFG system\footnote{\url{https://github.com/clulab/timenorm}} \citep{bethard13scfg} and uses the original English grammar
and our own grammar for Spanish.

\begin{figure*}
    \centering
    \includegraphics[width=0.65\textwidth]{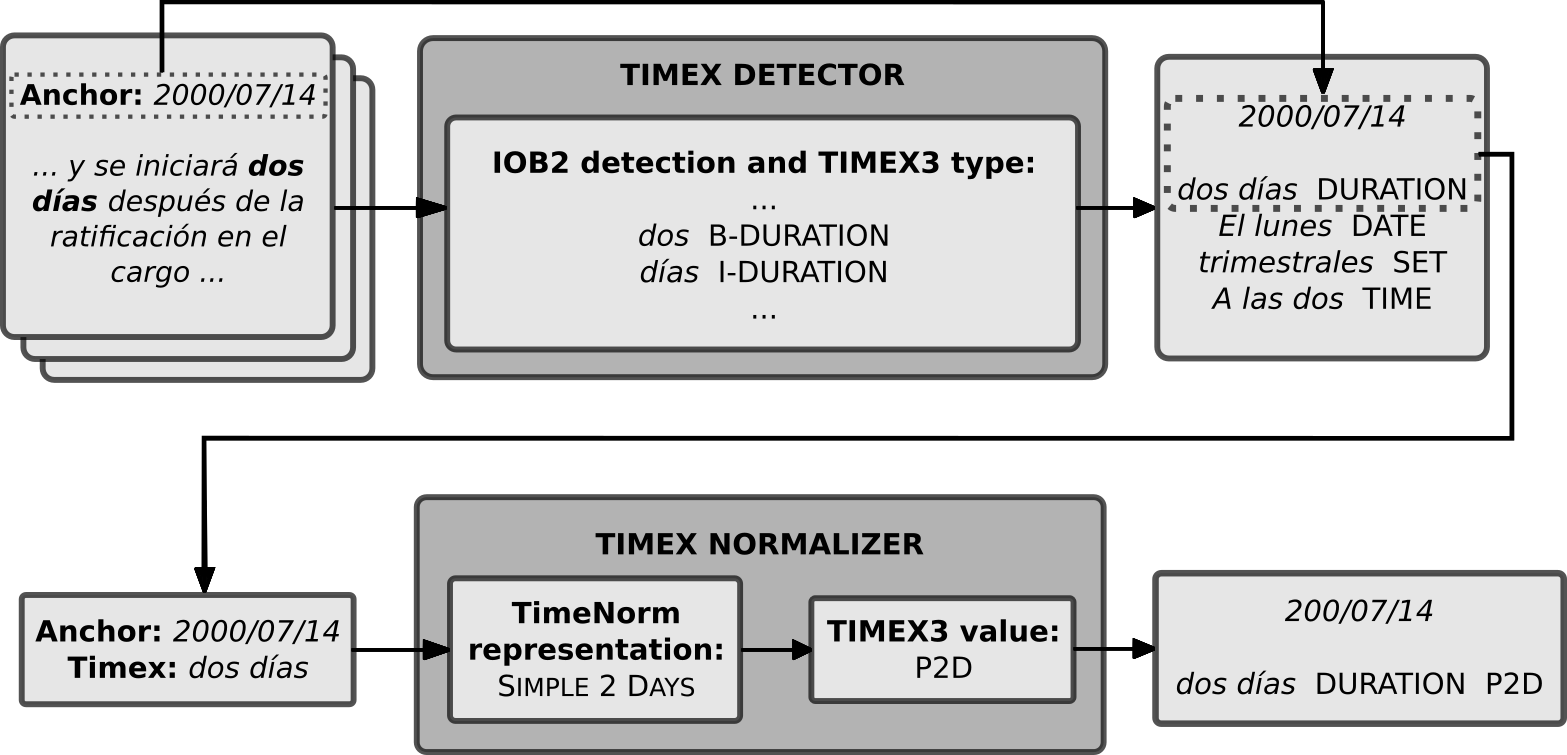}
    \caption{Overview of the system for detection and normalization.}
    \label{fig:system_overview}
\end{figure*}

Figure \ref{fig:system_overview} shows the complete process of timex detection and
normalization. For each document, our detector takes the document creation time (DCT) as anchor (e.g.
\textit{2000/07/14}), looks for possible time expressions (e.g. \textit{dos
días}, \textit{two days}) and tags the text token by token as belonging to one
of the possible timex types or not (e.g. ``DURATION''). Extracted timexes and
the corresponding anchor are sent to the normalizer, which parses each isolated timex
and generates its TimeNorm representation (e.g. ``\textsc{Simple 2 Days}'').
From this representation, the normalizer provides the appropriate value (e.g.
``P2D'').

\subsection{Timex Detection with XLM-RoBERTa}\label{ssec:detection}

Transformer-based multilingual masked language models such as XLM-RoBERTa have
proven solid cross-lingual performance in several tasks, such as natural
language inference, question answering and named entity recognition
\citep{devlin19bert,conneau20xlmroberta} thanks to the massive amounts of data
from many different languages that are used to train them.

XLM-RoBERTa is fine-tuned on the combination of English and Spanish TimeBank
training datasets. This combination performs better than fine-tuning only on
the separate monolingual datasets, and allows for timex detection on mixed multilingual documents. We found the best setting using the combined
training set also as development set for 5 epochs with 5 different seeds and a
learning rate of $5e^{-5}$. We tested other MLMs such as mBERT
\citep{devlin19bert} and IXAmBERT \citep{otegi20ixambert}, as well as Flair
contextual string embeddings \citep{akbik18contextemb}, but their performance
were substantially lower than that of XLM-RoBERTa.

\subsection{Timex Normalization with TimeNorm SCFG}

The system takes the candidate for normalization and its temporal anchor, which
must refer to a complete date or a complete date and time. These timexes are
tokenized to deal with different punctuation and segmentation cases. Although
the Italian TimeNorm \citep{mirza&minard15itTN} extends the English tokenizer to
manage some issues related to inflectional languages like Spanish, we preferred
to keep the default tokenizer for the Spanish version and address those issues
in the grammar. After tokenization, the timex is parsed on the TimeNorm synchronous
context-free grammar (SCFG) using an extended version of the CYK+ algorithm
\citep{chappelierCYK+}.

A SCFG is a kind of context-free grammar that builds two different trees
simultaneously to parse an expression: in our case, the source tree parses the
timex itself, whereas the target tree parses the corresponding TimeNorm temporal object. Non-terminals from both trees are aligned to keep track of the process, and can
be of different types: \textsc{[TimeSpan]} (e.g., corresponding to the source terminals \textit{la
noche de hace dos días, the night two days ago}), \textsc{[Period]} (e.g.
\textit{dos días, two days}), \textsc{[Field]} (e.g. \textit{noche, night}),
\textsc{[Unit]} (e.g. \textit{días, days}), \textsc{[Int]} (e.g. \textit{dos,
two}) or [\textsc{Nil}] (e.g. \textit{la, the}). On the contrary, each tree has its own kind of terminals: tokens in the source tree (e.g. \textit{marzo}, \textit{March}) and temporal operators in the target tree (e.g. \textsc{MonthOfYear} and \textsc{3}).

\begin{figure*}
\centering
\begin{subfigure}[Source tree]
    {\includegraphics[width=0.3\textwidth]{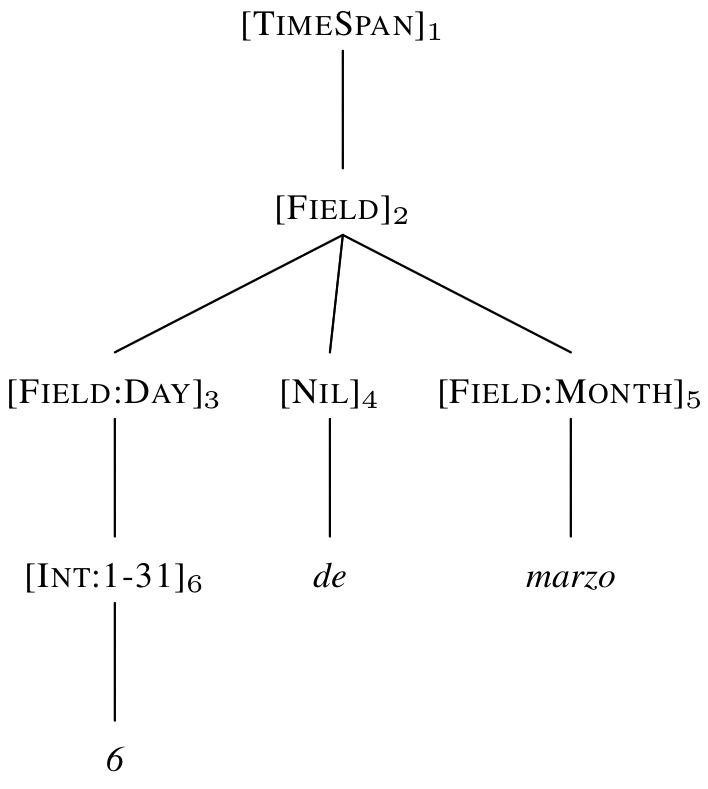}}
\end{subfigure}
\hfill
\begin{subfigure}[Target tree]
    {\includegraphics[width=0.6\textwidth]{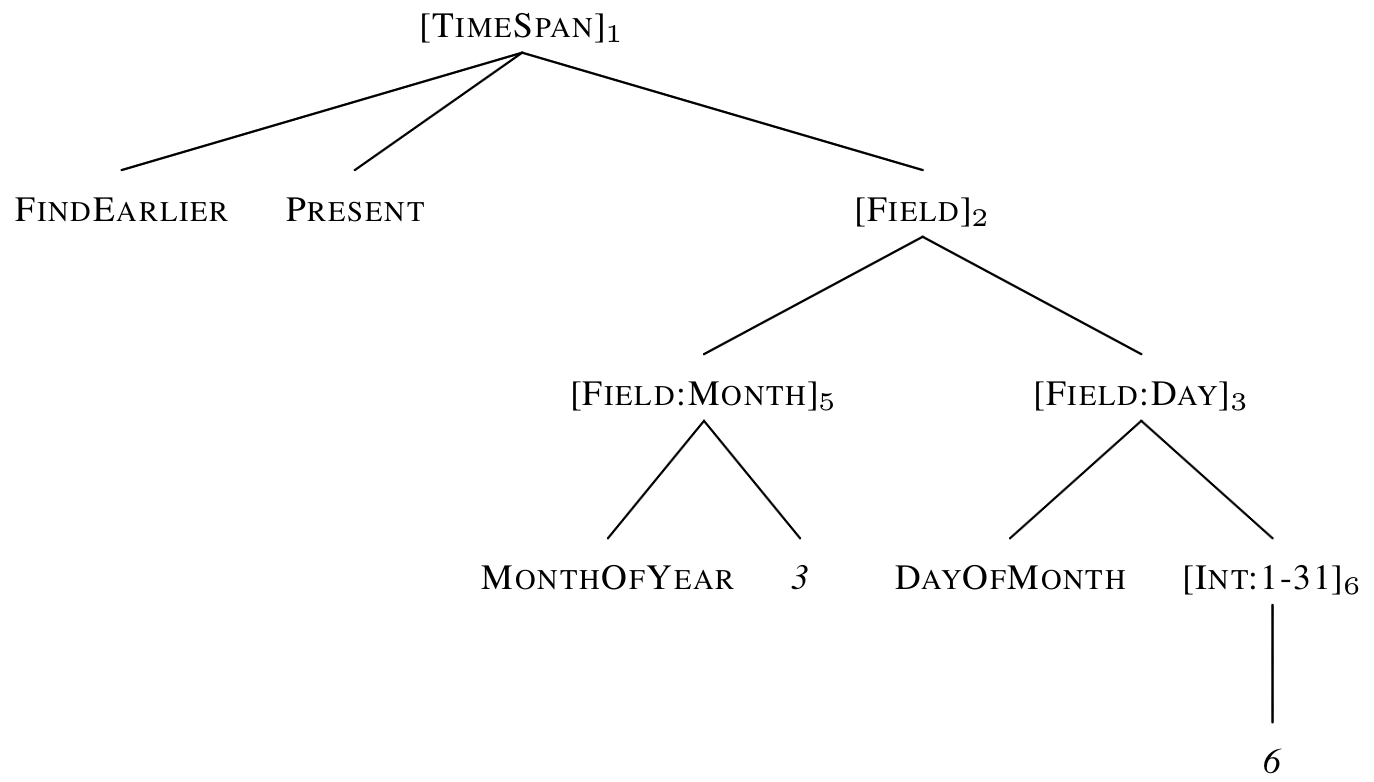}}
\end{subfigure}
\caption{TimeNorm SCFG parsing process for the timex \textit{6 de marzo}.}
\label{fig:trees}
\end{figure*}

\begin{table}
    \centering
    \scriptsize
    \begin{tabular}{r c l} \hline
        PARENT & $\rightarrow{}$ & SOURCE CHILD \\ & $\rightarrow{}$ & TARGET CHILD \\\hline
        {\textsc{[Int:1-31]$_{6}$}} & $\rightarrow{}$ & \textit{6} \\ & $\rightarrow{}$ & \textsc{6} \\\hline
        {\textsc{[Field:Month]$_{5}$}} & $\rightarrow{}$ & \textit{marzo} \\ & $\rightarrow{}$ & \textsc{MonthOfYear 3} \\\hline
        {\textsc{[Nil]$_{4}$}} & $\rightarrow{}$ & \textit{de} \\ & $\rightarrow{}$ & \O \\\hline
        {\textsc{[Field:Day]$_{3}$}} & $\rightarrow{}$ & \textsc{[Int:1-31]$_{6}$} \\ & $\rightarrow{}$ & \textsc{DayOfMonth [Int:1-31]$_{6}$} \\\hline
        {\textsc{[Field]}}$_{2}$ & $\rightarrow{}$ & \textsc{[Field:Day]$_{3}$ [Field:Month]$_{5}$} \\ & $\rightarrow{}$ & \textsc{[Field:Month]$_{5}$ [Field:Day]$_{3}$} \\\hline
        {\textsc{[TimeSpan]}$_{1}$} & $\rightarrow{}$ & \textsc{[Field]}$_{2}$ \\ & $\rightarrow{}$ & \textsc{FindEarlier Present [Field]$_{2}$} \\\hline
    \end{tabular}
    \caption{Simplified TimeNorm SCFG rules to parse \textit{6 de marzo}.}
   \label{tab:rules}
\end{table}

The combination of terminals and non-terminals in the target tree generates the temporal objects that correspond to the source tree's timexes,
following the procedures defined by the temporal operators. For example, our
grammar parses \textit{6 de marzo} (\textit{March 6th}) as the temporal object
\textsc{FindEarlier(Present, (MonthOfYear=3, DayOfMonth=6))}, where the
instantaneous operator \textsc{Present} is expanded to the enclosing day, e.g.
the DCT. Thus, the grammar would normalize this timex as the March 6th previous
to the anchor. This output representation is turned into
a TimeML normalization value regarding very simple conversion rules. In fact,
different TimeNorm representations can lead to the same TimeML annotation,
since they usually convey more meaning complexity than TimeML.

The parsing process for \textit{6 de marzo} is shown in Figure \ref{fig:trees}, where
bracketed elements are non-terminals, the rest are terminals and subscripts are
the alignments between non-terminals.

Furthermore, a simplification of the rules designed to create this parsing can be seen in Table \ref{tab:rules}. Each rule indicates the transition from a parent node to its children, which are divided in two rows: the upper row corresponds to the source representation (e.g. \textit{marzo}) and the lower row corresponds to the target representation (e. g. \textsc{MonthOfYear} 3). Following the rules and the subscripts, we can see how to generate the simultaneous trees in Figure \ref{fig:trees} from the source terminals \textit{6}, \textit{de} and \textit{marzo} to the \textsc{[TimeSpan]} root in the last row\footnote{The whole process of the TimeNorm system is deeply explained in \cite{bethard13scfg}.}.

\subsection{Building the Spanish Grammar}

We developed our own Spanish grammar based on the previous TimeNorm versions
for English \citep{bethard13scfg} and Italian \citep{mirza&minard15itTN} and the
gold timexes in the Spanish TempEval-3 training corpus.

Our grammar development was guided by three principles: (i) to make the
necessary changes in the source grammar whenever possible instead of adding
preprocessing steps (as it was done for Italian): e.g. \textit{del año}, \textit{of the year} vs.
\textit{dell'anno}, or \textit{dos mil catorce}, \textit{twenty fourteen} vs.
\textit{duemilaquattordici}; (ii) to prefer including the most common
expressions rather than trying to cover too many different spelling and
grammatical possibilities, in order to make our grammar simpler, more compact, 
and to avoid unnecessary ambiguities: e.g. we parse \textit{treinta y nueve},
\textit{thirty-nine} but not \textit{treintainueve}, which is orthographically
correct but not frequent; (iii) to build a linguistically coherent grammar
rather than a task-oriented one, thereby facilitating the adaptation of our grammar to
different domains and ensuring the explainability of the rules, namely, we avoided
adding rules that may improve performance but that are grounded on any
linguistic intuition.

These principles are meant to build a compact but general grammar for Spanish.
In order to adapt TimeNorm to other languages, different guidelines may be
needed, and some issues should be taken into account. For instance, some
languages might benefit from more preprocessing to tokenize expressions (like
Italian), or to deal with complex morphosyntactic features in fusional and
agglutinative languages, rather than doubling rules. Nevertheless, although
rule doubling extends significantly the grammar length, it might avoid
ambiguities among different parses. Languages with rather free syntactic order
like Spanish may also need more rules to address all possibilities (e.g.
\textit{la siguiente semana} and \textit{la semana siguiente}, \textit{the next
week}). Moreover, problems with cultural conceptions may arise, such as
temporal spans that do not strictly match TimeNorm thresholds and are difficult
to represent in a logical way (e.g. \textit{mediodía}, \textit{midday} may be
considered as an exact time, a specific time span or an underspecified time span in
Spanish).

On the other hand, ambiguity among different parses is a common problem to
grammars, and it may not always be adequately solved. The TimeNorm SCFG
provides heuristics to choose a parse over another, such as preferring earlier
time spans to later ones when both are possible (e.g. \textit{6 de marzo}), but
due to the lack of context knowledge it might lead to some errors. Other
TimeNorm limitations include the lack of implementation of some temporal units
(e.g. the notion of \textit{semestre}, \textit{semester}) and the assignation
of field values to concrete days, instead of being allowed to remain abstractly
(e.g. \textit{la noche}, \textit{the night} is always assigned to a concrete
day).

Despite these issues, the clear distinction between the architecture of
the system and the grammar makes it easy to adapt the rules to languages like
Spanish without modifying in any way the internal structure of TimeNorm. Additionally, our approach allows for timex processing on mixed multilingual documents by just implementing a language identification module before timex normalization, given that the detection model is multilingual and the normalization grammar is independent from detection patterns, unlike systems such as HeidelTime.

\section{Experimental Results}\label{sec:results}

We compare out method with HeidelTime \citep{stroetgen13heidelTE3} in three
evaluation settings: gold timex normalization, as in \citet{bethard13scfg},
timex detection, and the combination of timex detection and normalization
(TempEval-3 relaxed value F1) \citep{uzzaman13te3}.

\subsection{Datasets}\label{ssec:datasets}

Both English and Spanish TempEval-3 TimeBank training sets (``TE3 Train'') were
used together for fine-tuning XLM-RoBERTa. Then, the TempEval-3 test sets
(``TE3 Eval'') were used to report final results for each language. 
We used this setup because results were better when using the multilingual
training data (more details in Section \ref{ssec:det-norm-results}). 
Moreover, we also tested our system on the MEANTIME
corpus (``MT'') \citep{minard16meantime}, consisting on a parallel dataset
available in different languages that allows us to further evaluate our system
performance.  Table \ref{tab:datasets} shows the distribution of the processed
files, tokens and timexes in the corpora. Time expressions include DCTs, which
are as numerous as the number of files.

\begin{table}
    \centering
    \begin{tabular}{l l r r r}
        \noalign{\hrule height 0.5pt}
        Dataset & Lang & Files & Tokens & Timexes \\
        \noalign{\hrule height 0.5pt}
        \multirow{2}{4.5em}{TE3 Train}
        & en & 183 & 62,682 & 1,426 \\
        & es & 175 & 58,627 & 1,269 \\
        \hline
        \multirow{2}{4.2em}{TE3 Eval}
        & en & 20 & 7,014 & 158 \\
        & es & 35 & 9,916 & 234 \\
        \hline
        \multirow{2}{4.2em}{MT}
        & en & 120 & 14,101 & 604 \\
        & es & 120 & 15,973 & 598 \\
        \noalign{\hrule height 0.5pt}
    \end{tabular}
    \caption{Datasets used for training and evaluation.}
    \label{tab:datasets}
\end{table}

\subsection{Gold Timex Normalization}\label{ssec:norm-results}

Although we evaluated the normalization of gold timexes from TempEval-3 and
MEANTIME corpora according to \citet{bethard13scfg}, unlike in that work, we
did not test the normalization of the DCTs but only use them as temporal
anchors. To evaluate normalization we obtain the accuracy of a system at
normalizing the gold timexes to their gold values. Table \ref{tab:results-norm}
presents the normalization accuracy results of TimeNorm and HeidelTime for
English and Spanish in TempEval-3 Eval (``TE3 Acc'') and MEANTIME (``MT Acc'')
corpora. We can observe that overall TimeNorm outperforms HeidelTime, except
for Spanish on TempEval-3 Eval, where the two systems achieve the same
normalization accuracy. The best monolingual system to date, UWTime, obtains 82.6 at accuracy on English TempEval-3 test set according to \cite{lee14uwtime}.



\begin{table}
    \centering
    \begin{tabular}{l l r r}
        \hline
        Lang & System & TE3 Acc & MT Acc \\
        \hline
        \multirow{2}{1em}{en}
        & TimeNorm & \textbf{78.99} & \textbf{76.86} \\
        & HeidelTime & 74.64 & 75.41 \\
        \hline
        \multirow{2}{1em}{es}
        & TimeNorm & \textbf{80.40} & \textbf{72.57} \\
        & HeidelTime & \textbf{80.40} & 71.61 \\
        \hline
    \end{tabular}
    \caption{Results for normalization of gold timexes from TempEval-3 and MEANTIME corpora.}
    \label{tab:results-norm}
\end{table}

\subsection{Timex Detection and Normalization}\label{ssec:det-norm-results}

\begin{table*}
	\footnotesize
    \centering
	\begin{tabular}{lll@{\hspace{0.2cm}}c@{\hspace{0.2cm}}c@{\hspace{0.2cm}}c@{\hspace{0.2cm}}c@{\hspace{0.2cm}}c@{\hspace{0.2cm}}c@{\hspace{0.2cm}}c} \hline
        Dataset & Lang & System & Rel P & Rel R & Rel F1 & Rel TF1 & Rel VP & Rel VR & Rel VF1 \\
        \hline
        \multirow{6}{*}{TE3 Eval} & \multirow{3}{*}{en}
        & XTN-D & 93.48 & \textbf{93.48} & \textbf{93.48} & \textbf{89.86} & 76.81 & 76.81 & 76.81 \\
        & & XTN-N & \textbf{95.31} & 88.41 & 91.73 & 88.72 & 82.81 & 76.81 & \textbf{79.70} \\
        & & HeidelTime & 93.13 & 88.41 & 90.71 & 83.27 & - & - & 78.07 \\
        \cline{2-10}
        & \multirow{3}{*}{es}
        & XTN-D & 95.98 & \textbf{95.98} & \textbf{95.98} & \textbf{90.95} & 75.88 & 75.88 & 75.88 \\
        & & XTN-N & \textbf{96.74} & 89.45 & 92.95 & 89.30 & 82.07 & 75.88 & 78.85 \\
        & & HeidelTime & 96.02 & 84.92 & 90.13 & 87.47 & - & - & \textbf{85.33} \\
        \hline
        \multirow{6}{4em}{MT} & \multirow{3}{2em}{en}
        & XTN-D & 95.48 & \textbf{91.74} & \textbf{93.57} & \textbf{88.72} & 77.85 & 74.79 & 76.29 \\
        & & XTN-N & \textbf{96.37} & 87.81 & 91.89 & 87.14 & 82.31 & 75.00 & 78.49 \\
        & & HeidelTime & 94.91 & 84.71 & 89.52 & 85.15 & \textbf{84.49} & \textbf{75.41} & \textbf{79.69} \\
        \cline{2-10}
        & \multirow{3}{2em}{es}
        & XTN-D & 90.14 & \textbf{95.41} & 92.70 & 85.19 & 67.65 & 71.61 & 69.57 \\
        & & XTN-N & \textbf{91.99} & 93.53 & \textbf{92.75} & \textbf{85.51} & 70.64 & \textbf{71.82} & 71.22 \\
        & & HeidelTime & 89.79 & 88.10 & 88.94 & 81.35 & \textbf{72.98} & 71.61 & \textbf{72.29} \\
        \hline
    \end{tabular}
    \caption{Results for timex detection and normalization on TempEval-3 Eval and MEANTIME.}
    \label{tab:results-det-norm}
\end{table*}

In order to test our system on TempEval-3 metrics, we propose the evaluation of
timex extraction as a sequence labelling task: a strict match requires the
detection of exactly the tokens in a gold span, while a relaxed match only
needs the detection of a gold token. TIMEX3 types may be tagged as
sequence-labelling classes for each token (e.g. ``B-DURATION'' for
\textit{dos}, ``I-DURATION'' for \textit{días}), whereas TIMEX3 normalization
values should appear as additional information for all the tokens corresponding
to the gold or predicted timex (e.g. ``P2D'' for both \textit{dos} and
\textit{días}). These matches are computed to obtain the precision, recall and
F1 score for strict and relaxed detection, as well as for type and value on
correctly detected timexes, as specified in TempEval-3. 










It should be remembered that our detection XLM-RoBERTa-based models are fine-tuned
(using the parameters described in \ref{ssec:detection}) on mutilingual,
Spanish and English, training data. This is due to the fact that this setting
outperformed its monolingual counterpart (consisting of fine-tuning on the
target language only). Thus, for English the monolingual setting obtained an
strict detection F1 score of 87.27, substantially lower than the 88.28 F1 score
obtained when fine-tuning with multilingual training data. In Spanish the
scores were 88.12 for the monolingual setting with respect to the 88.28 strict
F1 score in the multilingual case.

As a modular system, our approach detects timexes without considering if they
can be easily normalized or not. On the contrary, in HeidelTime's integrated
approach its rules are designed to detect those that can normalize. For
example, the timex \textit{the day}, which may be normalized as any concrete or
underspecified day depending on the context, is detected by our system, but not by
HeidelTime. To analyze how this decision affects the results, we propose a
detection-focused approach (XTN-D), which detects timexes even if they may not
be normalized, and a normalization-focused approach (XTN-N), which ignores
timexes for which TimeNorm fails to provide a normalization (incorrect or not).
Still, and unlike Heideltime, TimeNorm will try to normalize complex cases such
as \textit{the day}, because it has been detected by the timex detector. 


To evaluate
Heideltime\footnote{\url{https://github.com/HeidelTime/heideltime/releases/tag/VERSION2.2.1}}
with the MEANTIME corpus, we fed HeidelTime the raw texts for the corresponding
language with their DCTs and set the domain to ``NEWS'', processed the output
and evaluated the predictions\footnote{The results obtained by HeidelTime
Standalone (version 2.2.1) on English TempEval-3 according to the official
evaluator and our own were slightly different from those published.}, as well
as our own results. Note that this method was also used to evaluate HeidelTime
gold timex normalization from Section \ref{ssec:norm-results}. To compare our
method in TempEval-3, we report the results published in their official
repository\footnote{\url{https://github.com/HeidelTime/heideltime/wiki/Evaluation-Results}}.
Table \ref{tab:results-det-norm} presents the relaxed detection precision
(``Rel P''), recall (``Rel R'') and F1 (``Rel F1''), type F1 (``Rel TF1'') and
value precision (``Rel VP''), recall (``Rel VR'') and F1 (``Rel VF1'') from
TempEval-3 metrics of XTN-D, XTN-N and HeidelTime on TempEval-3 and MEANTIME.
Results show that our multilingual fine-tuning method for timex detection is
substantially better than HeidelTime at detection and type recognition, whereas
HeidelTime obtains better scores the combined TempEval relaxed value F1 metric
(Rel VF1), which evaluates together normalization and detection, except in
English TempEval-3, where our method is slightly better. It is worth
highlighting the gap at Spanish TempEval-3 Rel VF1 between HeidelTime and our
systems, especially considering both the detection and the gold timex
evaluation presented in Table \ref{tab:results-norm}. As the best monolingual system to date, UWTime obtains 91.4 at Rel F1, 85.4 at Rel TF1 and 82.4 at VF1 on English TempEval-3 test set according to \cite{lee14uwtime}.

\section{Discussion}\label{sec:discussion}

\begin{figure*}
    \centering
    \includegraphics[width=0.85\textwidth]{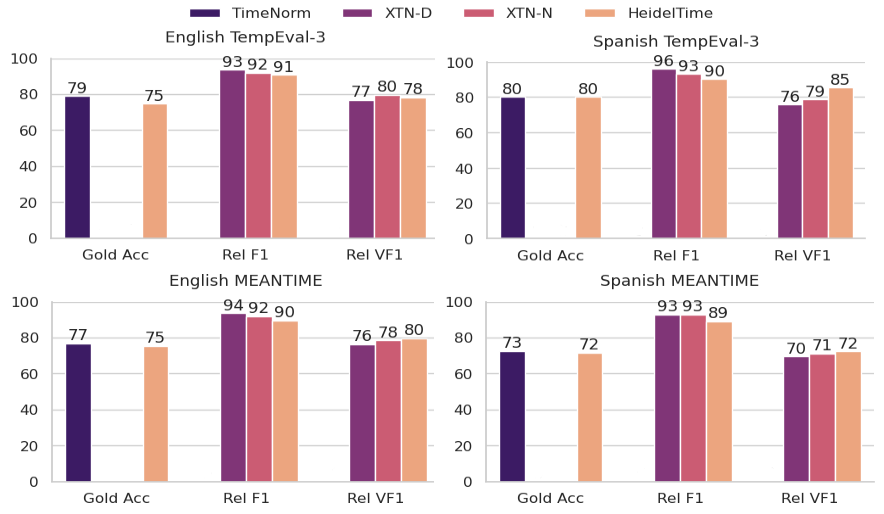}
    \caption{Comparison of gold normalization accuracy, relaxed detection F1 and relaxed value F1.}
    \label{fig:scores}
\end{figure*}

It might seem that there is a certain inconsistency in the results reported for
the gold timex normalization (Section \ref{ssec:norm-results}) and for the
TempEval-3 timex detection and normalization (Section
\ref{ssec:det-norm-results}). Thus, while our system achieves higher scores at
gold accuracy and relaxed detection F1, the combined relaxed value F1 scores
are in general lower than those obtained by HeidelTime. After a manual
inspection of the predictions, we observed that one of the reasons for this is
the aforementioned HeidelTime strategy to leave timexes that are difficult to
normalize undetected, raising the number of false negatives at detection. These
false negatives are computed as wrong values at gold timex normalization, while
according to TempEval-3 metrics they are considered as wrong span matches and
thus do not count as normalization errors. Since TempEval-3 value F1 depends on
the detection F1 and the number of correct values over the number of correct
detection matches, this specific metric favors the HeidelTime strategy. On the
contrary, our normalization-focused system XTN-N does not inform the
normalization process about the complexity of some timexes, and TimeNorm may
provide wrong values to these timexes without leaving them undetected. To
illustrate these differences among metrics, Figure \ref{fig:scores} shows a
comparison of TimeNorm and HeidelTime gold timex normalization accuracy (``Gold
Acc''), on the one hand, and relaxed detection F1 (``Rel F1'') and relaxed
value F1 (``Rel VF1'') for XTN-D, XTN-N and HeidelTime, on the other. In
these charts we can see how, even with higher or equal gold accuracy and
relaxed detection F1, our relaxed value F1 scores are lower than HeidelTime's
scores, except for English TempEval-3.

 \begin{table*}
      \centering
      \small{
     \begin{tabular}{l r r r r}
         \hline
         & \multicolumn{2}{c}{English} & \multicolumn{2}{c}{Spanish} \\
         \hline
         Error type & TimeNorm & HeidelTime & TimeNorm & HeidelTime \\
         \hline
         Overall & 29 & 35 & 39 & 39 \\
         \hline
         Not detected & - & 17 & - & 30 \\
         Wrong span & - & 5 & - & 5 \\
         Lack of context & 2 & 0 & 5 & 0 \\
         Lack of rules & 9 & 0 & 9 & 0 \\
         Wrong disambiguation & 5 & 3 & 13 & 0 \\
         Wrong underspecified time span & 6 & 3 & 3 & 0 \\
         Wrong value class & 6 & 4 & 8 & 3 \\
         Equivalent value & 1 & 3 & 1 & 1 \\
         \hline
     \end{tabular}
     }
     \caption{Errors made by TimeNorm and HeidelTime at normalizing TempEval-3 gold timexes.}
     \label{tab:errors-norm}
 \end{table*}

 \begin{table*}
     \centering
     \small{
     \begin{tabular}{@{\hspace{0.2cm}}l@{\hspace{0.2cm}} l@{\hspace{0.2cm}} r@{\hspace{0.2cm}} r@{\hspace{0.2cm}} r@{\hspace{0.2cm}} r@{\hspace{0.2cm}} r@{\hspace{0.2cm}} r}
         \hline
         & & \multicolumn{3}{c}{English} & \multicolumn{3}{c}{Spanish} \\
         \hline
         & Error type & XTN-D & XTN-N & HT & XTN-D & XTN-N & HT \\
         \hline
         Overall & Overall & 41 & 38 & 44 & 56 & 54 & 47 \\
         \hline
         \multirow{3}{3em}{Det} 
         & Overall - Det & 18 & 22 & 26 & 16 & 27 & 38 \\
         & False positives & 9 & 6 & 9 & 8 & 6 & 8 \\
         & False negatives & 9 & 16 & 17 & 8 & 21 & 30 \\
         \hline
         \multirow{7}{3em}{Norm} 
         & Overall - Norm & 23 & 16 & 18 & 40 & 27 & 9 \\
         & Wrong span & 1 & 0 & 5 & 3 & 1 & 5 \\
         & Lack of context & 1 & 0 & 0 & 3 & 0 & 0 \\
         & Lack of rules & 5 & 0 & 0 & 8 & 0 & 0 \\
         & Wrong disambiguation & 5 & 5 & 3 & 12 & 12 & 0 \\
         & Wrong underspecified time span & 6 & 6 & 3 & 3 & 3 & 0 \\
         & Wrong value class & 4 & 4 & 4 & 10 & 10 & 3 \\
         & Equivalent value & 1 & 1 & 3 & 1 & 1 & 1 \\
         \hline
     \end{tabular}
     }
     \caption{Errors made by each system on TempEval-3 timex detection and normalization task.}
     \label{tab:errors-det-norm}
 \end{table*}


In order to understand better the systems behaviour, we present the type of
errors made by our system and HeidelTime for gold timex normalization in Table
\ref{tab:errors-norm} and for combined timex detection and normalization in
Table \ref{tab:errors-det-norm}. Detection errors are classified as false
positives and false negatives, while normalization errors are presented in a
larger distribution depending on the source of the mistake:

\begin{itemize}
    \item A wrongly predicted span: e.g. prediction of \textit{15 días} (\textit{15 days}) $\rightarrow$ ``P15D'' instead of \textit{dentro de 15 días} (\textit{in 15 days}) $\rightarrow$ ``1999-06-17''.
     \item A lack of enough context for the system: e.g. no normalization for \textit{cinco} (\textit{five}) $\rightarrow$ ``P5M'' in \textit{suspendió con cuatro meses a Guardiola y con cinco a Stam} (\textit{they suspended Guardiola for four months and Stam for five}).
     \item A lack of rules that could match the timex: e.g. \textit{un momento dado de la historia} (\textit{a given moment in history}) $\rightarrow$ ``PAST\_REF''.
     \item Wrong temporal disambiguation: e.g. \textit{April 7} $\rightarrow$ ``2012-04-07'' instead of ``2013-04-07''.
     \item Wrong underspecified time span: e.g. \textit{ese día} (\textit{that day}) $\rightarrow$ ``2002-02-01'' instead of ``2002-XX-XX''.
     \item Wrong value class: e.g. \textit{un año} (\textit{a year}) $\rightarrow$ the year period ``P1Y'' instead of the concrete year ``1999''.
     \item Equivalent value: e.g. \textit{a decade} $\rightarrow$ ``P1DE'' instead of ``P10Y''.
\end{itemize}








Table \ref{tab:errors-norm} shows that some of their errors at gold timex
normalization stem from detection, such that timexes are not detected or the
system tries to normalize a wrong span. These mistakes, which are the most
numerous in HeidelTime (22 in English, 35 in Spanish), cannot be produced by
TimeNorm. Contrarily, TimeNorm makes more grammatical mistakes (29 in English,
39 in Spanish) than HeidelTime (13 in English, 4 in Spanish), being wrong
disambiguations, lack of rules, wrong value class and wrong underspecified time span
the main sources of error. Nevertheless, it must be noted that, in the overall
results on TempEval-3, TimeNorm makes less mistakes in English and the same number in
Spanish, beating or matching HeidelTime at gold timex normalization.  With
respect to timex detection and normalization errors on TempEval-3 in Table
\ref{tab:errors-det-norm}, our method makes less mistakes than HeidelTime at
detection, but more at normalization except for XTN-N in English. In fact, the
XTN-N obtains better results than XTN-D by reducing the number of normalization
errors in exchange of more false negatives detection errors. It should also be
noted the large amount of false negatives produced by HeidelTime in Spanish
compared to the number of normalization errors, as shown by Table
\ref{tab:errors-norm}. In particular, Spanish HeidelTime does not make any
wrong disambiguation, whereas this is the biggest problem for Spanish TimeNorm
in this corpus, followed by assigning a wrong value class or underspecified time
span (the lack of rules only affects XTN-D).

The results and error analysis suggest that an end-to-end strategy to decide if
timexes can be normalized, and to ignore them if not, is highly beneficial when
evaluating using the TempEval-3 relaxed value F1 metric. However, this is not
the case when evaluating in the gold timex normalization or timex detection
settings. Our analysis therefore raises the question as to what constitutes the
best strategy for timex processing. Is it more convenient to ignore timexes for
which it is not easy to provide a correct normalization as Heideltime does?
Should we try to detect and normalize them to obtain a better coverage, as our
method tries to do?

Thus, our modular approach combines a multilingual MLM and a grammar that
obtains state-of-the-art results in gold timex normalization, timex detection
and type recognition, which can be easily adapted to other languages. However,
it lacks an end-to-end strategy to ignore difficult timexes and suffers from
lack of context for some complex disambiguations. 

\section{Concluding Remarks}\label{sec:conclusions}

In this paper we present a modular system for the multilingual detection and
normalization of temporal expressions or timexes, as defined in TempEval-3. Our
system consists of a Transformer-based multilingual neural detector based on
XLM-RoBERTa and a rule-based normalizer on the TimeNorm SCFG architecture. This
modularity allows us to combine the best approaches for each of the two tasks,
achieve state-of-the-art results and facilitate the adaptation to other
languages, due to the multilingual training of the neural detector and the easy
process to build such a normalization grammar.

In-domain and out-of-domain evaluation of our system for English and Spanish
shows better performance at gold timex normalization, timex detection and type
recognition than HeidelTime. However, in the combined detection and
normalization task (TempEval-3 relaxed value F1 metric), our system obtains
similar results to HeidelTime except for the in-domain Spanish evaluation.
Results indicate that an end-to-end strategy to ignore those timexes that are
difficult to normalize is highly beneficial in the TempEval-3 evaluation
setting. 

It should be noted that in this paper we only address detection and normalization of temporal timexes
in two languages, namely, English and Spanish. Furthermore, previous work on
these tasks is not very recent, with the state-of-the-art results for temporal
detection and normalization having been published several years ago
\citep{bethard13scfg,stroetgen15heidelmulti}. However, this is a reflection of
the state of the task itself. While there is some recent work on temporal
detection only \citep{lange20multi} no recent approach has addressed both
tasks, detection and normalization, as defined in the TempEval-3 shared task
\citep{uzzaman13te3}. Thus, this paper aims to address this particular
shortcoming while showing that further work is required on the combined task of
temporal detection and normalization.

In any case, we believe that our main conclusions still hold. Apart from
creating the first TimeNorm grammar for normalization in Spanish, we also
present the first modular method combining both grammar-based and deep learning
methods for both temporal normalization and detection. The results obtained
show that when evaluating on normalization, our approach outperforms the
state-of-the-art. Furthermore, while our detection results are also better than
previous work, the joined TempEval-3 evaluation method favours systems
with lower recall. Our detailed analysis shows that our modular
approach makes a similar or inferior number of errors to HeidelTime. However,
this is not reflected in the final F1 combined scores.

\section*{Acknowledgements}

This work has been supported by the HiTZ center and the Basque Government
(Research group funding IT-1805-22). Nayla Escribano is funded by a
``Formación de Personal Investigador'' grant from the Basque Government. We also acknowledge the funding from
the following MCIN/AEI/10.13039/501100011033 projects: (i) DeepKnowledge (PID2021-127777OB-C21)
and ERDF A way of making Europe; (ii) Disargue (TED2021-130810B-C21) and European Union
NextGenerationEU/PRTR (iii) Antidote (PCI2020-120717-2) and European Union NextGenerationEU/PRTR.
Rodrigo Agerri currently holds the RYC-2017-23647 fellowship
(MCIN/AEI/10.13039/501100011033 and ESF Investing in your future).

\bibliography{references}

\end{document}